%% file: visualie.tex
\documentclass{article}
\pdfoutput=1

\usepackage[ruled]{algorithm2e}
\usepackage{graphicx}
\usepackage{authblk}
\newtheorem{example}{Example}[section]
\usepackage{multirow}
\usepackage{graphicx}
\usepackage[below]{placeins}
\usepackage{url}
\usepackage[all=normal,floats=tight,leading=tight,%
paragraphs=tight,charwidths=tight,tracking=tight,%
wordspacing=tight]{savetrees}

\SetAlFnt{\small}
\SetAlCapFnt{\small}
\SetAlCapNameFnt{\small}
\SetAlCapHSkip{0pt}
\IncMargin{-\parindent}

\begin{document}
\linespread{0.8}
\setlength{\textfloatsep}{1pt}


\title{WYSIWYE\thanks{What You See Is What You Extract}: An Algebra for Expressing Spatial and Textual Rules for Visual Information Extraction}






\author[1] {Vijil Chenthamarakshan\thanks{ecvijil@us.ibm.com}}
\author[1] {Prasad .M. Deshpande\thanks{prasdesh@in.ibm.com}}
\author[1] {Raghu Krishnapuram\thanks{kraghura@in.ibm.com}}
\author[2] {Ramakrishna Varadarajan\thanks{ramkris@cs.wisc.edu}}
\author[3] {Knut Stolze\thanks{ramkris@cs.wisc.edu}}
\affil[1]{IBM Research}
\affil[2]{University of Wisconsin-Madison}
\affil[3]{IBM Germany Research \& Development}
\date{}

\maketitle              
\input{abstract.tex}
\input{intro.tex}

\input{related-work.tex}

\input{visual-algebra}

\input{SQL-implementation}

\input{experiments}

\input{conclusions}
\bibliographystyle{abbrv}
\bibliography{sigproc}


%
%
%
%
%
%

\end{document}

%% file: abstract.tex
\begin{abstract}
The visual layout of a webpage can provide valuable clues for certain types of Information Extraction (IE) tasks. In traditional rule based IE frameworks, these layout cues are mapped to rules that operate on the HTML source of the webpages. In contrast, we have developed a framework in which the rules can be specified directly at the layout level.  This has many advantages, since the higher level of abstraction leads to simpler extraction rules that are largely independent of the source code of the page, and, therefore, more robust. It can also enable specification of new types of rules that are not otherwise possible. 
To the best of our knowledge, there is no general framework that allows declarative specification of information extraction rules based on spatial layout. Our framework is complementary to traditional text based rules framework and allows a seamless combination of spatial layout based rules with traditional text based rules. We describe the algebra that enables such a system and its efficient implementation using standard relational and text indexing features of a relational database. We demonstrate the simplicity and efficiency of this system 
for a task involving the extraction of software system requirements from software product pages.
\end{abstract}

%% file: intro.tex
\section{Introduction}
\label{sec:intro}
\begin{figure*}[ht]
	\centering
		\includegraphics[width=0.7\textwidth,height=0.7\textheight,keepaspectratio]{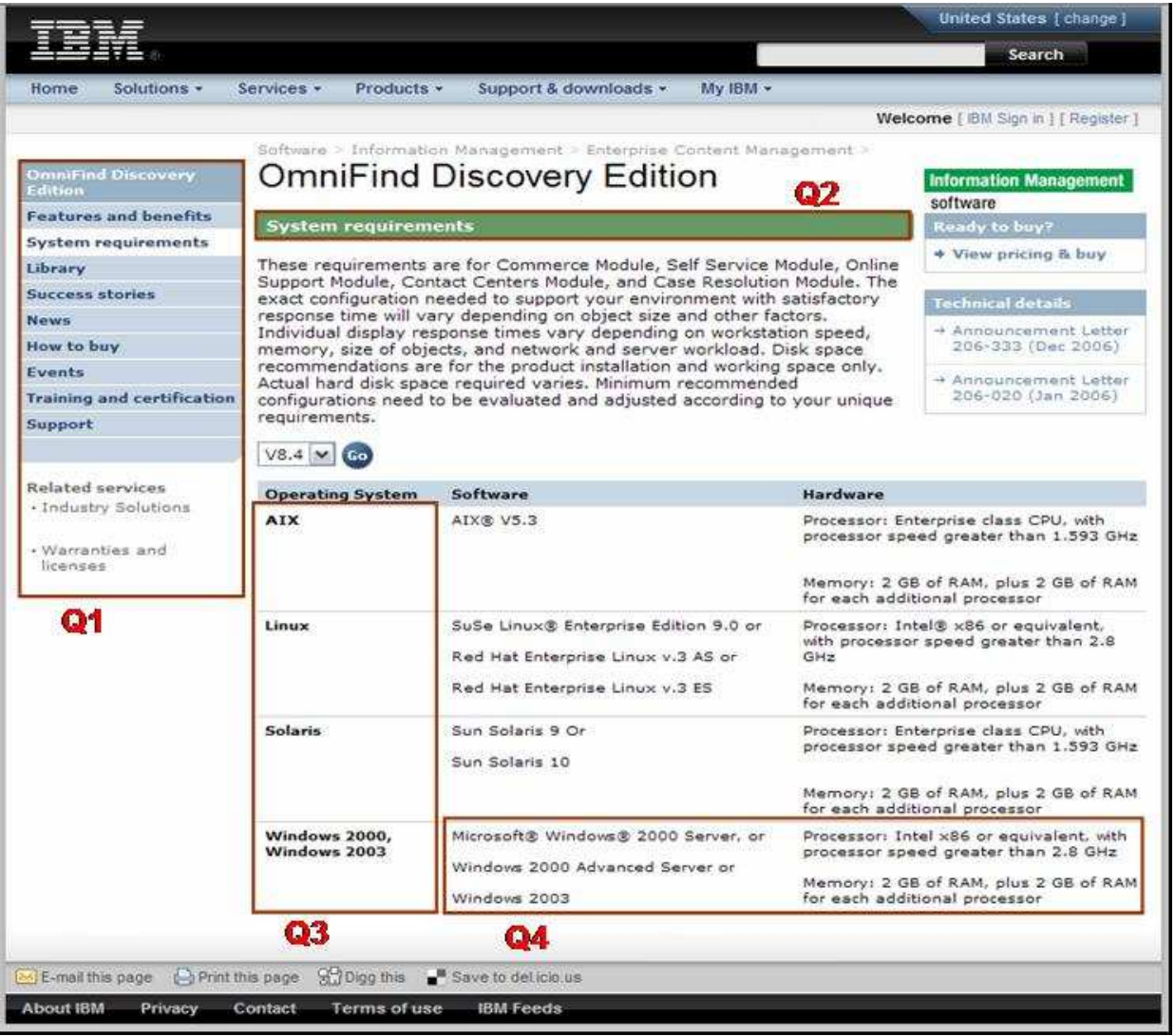}
	\caption{System requirements page}
	\label{fig:sys-req-2}
\end{figure*}

Information in web pages is laid out in a way that aids human perception using specification languages that can be understood by a web browser, such as HTML, CSS, and Javascript. 
The visual layout of elements in a page contain valuable clues that can be used for extracting information from the page. Indeed, there have been several efforts to use layout information for specific tasks such as web page segmentation~\cite{VIPS03} and table extraction~\cite{gatterbauer_towards_2007}. There are two ways to use layout information:
\begin{enumerate}
\item {\bf Source Based Approach:} Map the layout rule to equivalent rules based on the source code (html) of the page. For example, alignment of elements can be achieved in HTML by using a list ({\tt <li>}) or a table row ({\tt <tr>}) tag.
\item {\bf Layout Based Approach:} 
Use the layout information (coordinates) of various elements obtained by rendering the page to extract relevant information.
\end{enumerate}
Both these approaches achieve the same end result, but the implementations are different as illustrated in the example below.
\begin{example}
Figure~\ref{fig:sys-req-2} shows the system requirements page for an IBM software product. The IE 
task is to extract the set of operating systems supported by the product (listed in a column in the table indicated by Q3). In the source based approach, the rules need to identify the table, its rows and columns, the row or column containing the word `Operating Systems', and finally a list of entities, all based on the tags that can be used to implement them. In the layout based approach, the rule can be stated as: `\textit{From each System Requirements page, extract a list of operating system names that appear strictly\footnote{See section~\ref{sec:visual-operators} for a definition of strictness} to the south of the word `Operating Systems' and are vertically aligned}'. The higher level layout based rule is simpler, and is more robust to future changes in these web pages. 
\end{example}



%
%

Source based rules have several serious limitations, as listed below:
\begin{itemize}
	\item An abstract visual pattern can be implemented in many different ways by the web designer. For example, a tabular structure can be implemented using any of {\tt <table>}, {\tt <div>} and {\tt <li>} tags. Lerman et al~\cite{lerman04} show that only a fraction of tables are implemented using the {\tt <table>} tag. Source-based rules that use layout cues need to cover all possible ways in which the layout can be achieved. 
	Our experience with large scale IE 
	tasks suggest that rules that depend on HTML tags and DOM trees work reasonably well on template based machine--generated pages, but become too complex and brittle to maintain when applied to manually authored web pages.  
	\item Proximity of two entities in the HTML source code does not necessarily imply visual proximity~\cite{krpl_using_2005}, and so it may not be possible to encode visual proximity cues using simple source based rules.
	\item Specification languages are becoming more complex and difficult to analyze. Visualization logic is often embedded in CSS and Javascript, making the process of rule writing difficult. 
	\item Rules based on HTML tags and DOM trees are often sensitive to even minor modifications of the web page,
	and rule maintenance becomes messy. 
\end{itemize}

Layout based approaches overcome these limitations since they are at a higher level and independent of the page source code. Previous efforts at using layout based approaches were targeted at specific tasks such as page segmentation, wrapper extraction, table extraction, etc and are implemented using custom code. Existing rule based information extraction frameworks do not provide a mechanism to express rules based on the visual layout of a page.  Our goal is to address this gap by augmenting a rule based information extraction framework to be able to express layout based rules. Rule based system can be either declarative~\cite{AlgebraicIE,shen_declarative_2007} or procedural~\cite{patternSPL}. It has been shown that expressing information extraction (IE) tasks using an algebra, rather than procedural rules or custom code, enables systematic optimizations making the extraction process very efficient~\cite{AlgebraicIE,shen_declarative_2007}. 
Hence, we focus on an algebraic information extraction framework described in~\cite{AlgebraicIE} and extend its algebra with a visual operator algebra that can express rules based on spatial layout cues.
One of the challenges is that not all rules can be expressed using layout cues alone. For some rules, it may be necessary to use traditional text--based matching such as regular expressions and dictionaries, and combine them with spatial layout based rules. The framework thus needs to support rules that use both traditional textual matching and high--level spatial layout cues. 
In  summary, our contributions are as follows:
\begin{itemize}
\item We have developed an algebraic framework for rule--based information extraction that allows us to seamlessly combine traditional text--based rules with high--level rules based on the spatial layout of the page by extending an existing algebra for traditional text based information extraction ~\cite{AlgebraicIE}, with a visual operator algebra. 
We would like to reiterate that our focus is not on developing spatial rules for a specific task, rather we want to develop an algebra using which spatial rules for many different tasks can be expressed.
\item We implement the system using a relational database and demonstrate how the algebra enables optimizations by systematically mapping the algebra expressions to SQL. Thus, the system can benefit from the indexing and optimization features provided by relational databases.
\item We demonstrate the simplicity of the visual rules compared to source based rules for the tasks we considered. We also conduct performance studies on a dataset with about 20 million regions and describe our experience with the optimizations using region and text indices.   
\end{itemize}


%% file: related-work.tex
\section{Related Work}
\label{sec:related-work}
\textbf{Information Extraction(IE):} 
IE is a mature area of research that has received widespread attention in the NLP, AI, web and database communities~\cite{ieSurvey08}. Both rule based and machine learning based approaches have been proposed and widely used in real life settings. 
In this paper, we extend the operator algebra of System T~\cite{AlgebraicIE} to support rules based on spatial layout.

\textbf{Frameworks for Information Extraction:} The NLP community has developed several software architectures for sharing annotators, such as GATE~\cite{Cunningham96gate}
and UIMA~\cite{UIMAcite}. The motivation is to provide a reusable framework where annotators developed by different providers can be integrated and executed in a workflow. 



\textbf{Visual Information Extraction:} There is a lot of work on using visual information for specific tasks. We list some representative work below. The VIPS algorithm described in~\cite{VIPS03} segments a DOM tree based on visual cues retrieved from browser's rendering. The VIPS algorithm complements our work as it can act as a good preprocessing tool performing task-independent page structure analysis before the actual visual extraction takes place - thereby improving extraction accuracy.  A top-down approach to segment a web page and detect its content structure by dividing and merging blocks is given in~\cite{VBWA}.  \cite{Mtree02} use visual information to build up a ``M-tree'', a concept similar to the DOM tree enhanced with screen coordinates.  \cite{gatterbauer_towards_2007} describe a completely domain-independent method for IE from web tables, using visual information from Mozilla browser. 
All these approaches are implemented as monolithic programs that are meant for specific tasks. On the other hand, we are not targeting a specific task; rather our framework can be used for different tasks by allowing declarative specification of both textual and visual extraction rules.

Another body of work that is somewhat related is automatic and semi-automatic wrapper induction for information extraction ~\cite{DBLP:conf/sigmod/ArasuG03}.
These methods learn the a template expression for extracting information based on some training sets. The wrapper based methods work well on pages that have been generated using a template, but do not work well on human authored pages. 

%% file: visual-algebra.tex
\section{Visual Algebra}
\label{sec:visual-algebra}

\subsection{Overview of Algebraic Information Extraction}
We start with a system proposed by Reiss et al~\cite{AlgebraicIE} and extend it to support visual extraction rules. First, we give a quick summary of their algebra. For complete details, we request the reader to refer to the original paper.
\subsubsection*{Data Model}
A document is considered to be a sequence of characters ignoring its layout and other visual information. The fundamental concept in the algebra is that of a span, an ordered pair $\left\langle begin, end \right\rangle$ that denotes a region or text within a document identified by its ``begin'' and ``end'' positions. Each annotator finds regions of the document that satisfy a set of rules, and marks each region with an object called a span. 

The algebra operates over a simple relational data model
with three data types: span, tuple, and relation. A tuple is an finite sequence of $w$ spans $\langle s_1, ..., s_w \rangle$;
where $w$ is the width of the tuple. A relation is a multiset
of tuples, with the constraint that every tuple in the relation
must be of the same width. Each operator takes
zero or more relations as input and produces a single output
relation. 
\subsubsection*{Operator Algebra}
The set of operators in the algebra can be categorized
broadly into relational operators, span extraction operators,
and span aggregation operators as shown in Table~~\ref{table:operators}. Relational operators include the standard operators such as select,
project, join, union, etc. The span extraction operators identify segments of text that match some pattern and produce spans corresponding to these matches. The two common span extraction operators are the regular expression matcher $\epsilon_{re}$ and the dictionary matcher $\epsilon_d$. The regular expression matcher takes a regular expression $r$, matches it to the input text and outputs spans corresponding to these matches. The dictionary matcher takes a dictionary $dict$, consisting of a set of words/phrases, matches these to the input text and outputs spans corresponding to each occurence of a dictionary item in the input text. 

%

\begin{table}[tbh]
		\begin{tabular}{|l|l|l|}
		  \hline
		  Operator class & Operators & Explanation\\
		  \hline
			Relational & $\sigma, \pi, \times, \cup, \cap, \ldots$ & \\
			Span extraction& $\epsilon_{re}, \epsilon_d$ & \\
			Span aggregation& $\Omega_o, \Omega_c, \beta$ & \\
			\hline
			& $s_1 \preceq_d s_2$ & $s_1$ and $s_2$ do not overlap, $s_1$ precedes $s_2$,\\ 
			& &   and there are at most $d$ characters between them\\
			Predicates & $s_1 \simeq s_2$ & the spans overlap \\
			& $s_1 \subset s_2$ & $s_1$ is strictly contained within $s_2$ \\
			& $s_1 = s_2$ & spans are identical \\			
			\hline
		\end{tabular}
	\caption{Operator Algebra for Information Extraction}
	\label{table:operators}
\end{table}

\begin{table}[!tb]
	{\scriptsize
		\begin{tabular}{|p{2cm}|l|p{6cm}|}
		  \hline
		  & Operator & Explanation \\
		  \hline \hline
			\multirow{3}{2cm}{Span Generating} & $\Re(d)$ & Return all the visual spans for the document $d$   \\ 						
			& Ancestors($vs$) & Return all ancestor visual spans of $vs$ \\
		  & Descendants($vs$) & Return all descendant visual spans of $vs$  \\
		  \hline \hline
		  \multirow{2}{2cm}{Directional Predicate} & NorthOf($vs_1$, $vs_2$) & Span $vs_1$ occurs above $vs_2$ in the page layout  \\ 
			& StrictNorthOf($vs_1$, $vs_2$) & Span $vs_1$ occurs strictly above $vs_2$ in the page \\ 			
			\hline \hline
			\multirow{3}{2cm}{Containment Predicate}  & Contains($vs_1$, $vs_2$) & $vs_1$ is contained within $vs_2$ \\ 			
			& Touches($vs_1$, $vs_2$) & $vs_1$ touches $vs_2$ on one of the four edges \\
			& Intersects($vs_1$, $vs_2$) & $vs_1$ and $vs_2$ intersect \\
			\hline \hline
			\multirow{2}{2cm}{Generalization, Specialization} & MaximalRegion($vs$)/ & Returns the largest/smallest visual span $vs_m$ \\
			& MinimalRegion($vs$)  & that contains $vs$ and the same text content as $vs$\\ 						
			\hline \hline
			\multirow{4}{2cm}{Geometric} & Area($vs$) & Returns the area corresponding to $vs$ \\
			& Centroid($vs$) & Returns a visual span that has $x$ and $y$ \\
			& & coordinates corresponding to the centroid  \\
			& & of $vs$ and text span identical to $vs$  \\ 						
			\hline \hline
			\multirow{2}{2cm}{Grouping} & (Horizontally/Vertically)Aligned & Returns groups of horizontally/vertically aligned \\
			
			& ($VS$, consecutive, maxdist) & visual spans from $VS$. If the $consecutive$ flag is set, 
			 the visual spans have to be consecutive with no 
			 non-aligned span in between. The $maxdist$ limits 
			 the maximum distance possible between two  
			 consecutive visual spans in a group  \\
			\hline \hline
			\multirow{2}{2cm}{Aggregation} & MinimalSuperRegion($VS$) & Returns the smallest visual span that contains all the 
			 visual spans in set $VS$ \\
			& MinimalBoundingRegion($VS$) & Returns a minimum bounding rectangle of all 
			 visual spans in set $VS$ \\
			\hline
		\end{tabular}
		}
	\caption{Visual Operators}
	\label{table:visual-operators}
\end{table}

The span aggregation operators take in a set of input spans and produce a set of output spans by performing certain aggregate operations over the input spans. There are two main types of aggregation operators - $consolidation$ and $block$. The consolidation operators are used to resolve overlapping matches of the same concept in the text. Consolidation can be done using different rules. Containment consolidation ($\Omega_c$) is used to discard annotation spans that are wholly contained within other spans. Overlap consolidation ($\Omega_o$) is used to produce new spans by merging overlapping spans. The block aggregation operator ($\beta$) identifies spans of text enclosing a minimum number of input spans such that no two consecutive spans are more than a specified distance apart. It is useful in combining a set of consecutive input spans into bigger spans that represent aggregate concepts. The algebra also includes some new selection predicates that apply only to spans as shown in Table~\ref{table:operators}. 

\subsection{Extensions for Visual Information Extraction}
\label{section:algebra-extensions}
We extend the algebra described in order to support information extraction based on visual rules. In addition to the span, we add two new types in our model -- $Region$ and $VisualSpan$. A $Region$ represents a visual box in the layout of the page and has the attributes: $\langle x_l, y_l, x_h, y_h \rangle$. $(x_l, y_l)$ and $(x_h, y_h)$ denote the bounding box of the identified region in the visual layout of the document. We assume that the regions are rectangles, which applies to most markup languages such as HTML. 
A $VisualSpan$ is a combination of a text based span and a visual region with the following attributes: $\langle s, r \rangle$, where $s$ is a text span having attributes $begin$ and $end$ as before and $r$ is the region corresponding to the span. 

The operators are also modified to work with visual spans. The relational operators are unchanged. The span extraction operators are modified to return  visual spans rather than spans. For example, the regular expression operator $\epsilon_{re}$ matches the regular expression $r$ to the input text and for each matching text span $s$ it returns its corresponding visual span. 
Similarly, the dictionary matcher $\epsilon_d$ outputs visual spans corresponding to occurences of dictionary items in the input text. The behavior of the span aggregation operators ($\Omega_c$ and $\Omega_o$) is also affected. 
Thus containment consolidation $\Omega_c$ will discard visual spans whose region and span are both contained in the region and span of some other visual span. 
Overlap consolidation ($\Omega_o$) aggregates visual spans whose text spans overlap. It produces a new visual span whose text span is the merge of the overlapping text spans and bounding box is the region corresponding to the closest HTML element that contains the merged text span. 

There are two flavors to the block aggregation operator ($\beta$). The text block operator ($\beta_s$) is identical to the earlier $\beta$ operator. It identifies spans of text enclosing a minimum number of input spans such that no two consecutive spans are more than a specified distance apart. The region block operator ($\beta_v$) takes as input a X distance $x$ and Y distance $y$. It finds visual spans whose region contains a minimum number of input visual spans that can be ordered such that the X distance between two consecutive spans is less than $x$ and the Y distance is less than $y$. The text span of the output visual spans is the actual span of the text corresponding to its region.

The predicates described in Table~\ref{table:operators} can still be applied 
to the text span part of the visual spans. To compare the region part of the visual spans, we need many new predicates, which are described in the next section.
\subsection{Visual Operators}
\label{sec:visual-operators}
We introduce many new operators in the algebra to enable writing of rules based on visual regions. The operators can be classified as span generating, scalar or grouping operators and a subset has been listed in Table~\ref{table:visual-operators}. Many of these operators are borrowed from spatial (GIS) databases. For example, the operators Contains, Touches and Intersects are available in a GIS database like DB2 Spatial Extender\footnote{\url{http://www-01.ibm.com/software/data/spatial/db2spatial/}}. However, to our best knowledge this is the first application of using these constructs for Information Extraction. 

\subsubsection*{Span Generating Operators}
These operators produce a set of visual spans as output and include the $\Re(d)$, $Ancestors$ and $Descendents$ operators. 

\subsubsection*{Scalar Operators} 
The scalar operators take as input one or more values from a single tuple and return a single value. 
Boolean scalar operators can be used in predicates and are further classified as directional or containment operators. The directional operators allow visual spans to be compared based on their positions in the layout. Due to lack of space, we have listed only $NorthOf$, however we have similar predicates for other directions.   
Other scalar operators include the generalization/specialization operators and the geometric operators.

\subsubsection*{Grouping Operators}
The grouping operators are used to group multiple tuples based on some criteria and apply an aggregation function to each group, similar to the GROUP BY functionality in SQL. 

%


\subsection{Comparison with Source Based Approach}
\label{sec:comparison-source}
If visual algebra is not supported, we would have to impelement a given task using only source based rules. The visual algebra is a superset of the existing source based algebra. Expressing a visual rule using existing algebra as a source based rule can be categorized into one of the following cases:
\begin{enumerate}
\item{\bf Identical Semantics:} Some of the visual operators can be mapped directly into source level rules keeping the semantics intact. For example, the operator $Vertically\-Aligned$ can be mapped to an expression based on constructs in html that are used for alignment such as {\tt <tr>}, {\tt <li>} or {\tt <p>}, depending on the exact task at hand. 
\item{\bf Approximate Semantics:} Mapping a visual rule to a source based rule with identical semantics may lead to very complex rules since there are many ways to achieve the same visual layout. 
It may be possible to get approximately similar results by simplifing the rules if we know that the layout for the pages in the dataset is achieved in one particular way. 
For example, in a particular template, alignment may always be implemented using rows of a table (the {\tt <tr>} tag), so the source based rule can cover only this case. 
\item{\bf Alternate Semantics:} In some cases, it is not possible to obtain even similar semantics from the source based rules. For example, rules based on $Area$, $Centroid$, $Contains$, $Touches$ and $Intersects$ cannot be mapped to source based rules, since it is not possible to check these conditions without actually rendering of the page. In such cases, we have to use alternate source based rules for the same task. 
\end{enumerate}

%% file: SQL-implementation.tex
\section{System Architecture and Implementation
}
\label{sec:sys-architecture}
\begin{figure*}[bt]
	\centering
		\includegraphics[width=8cm]{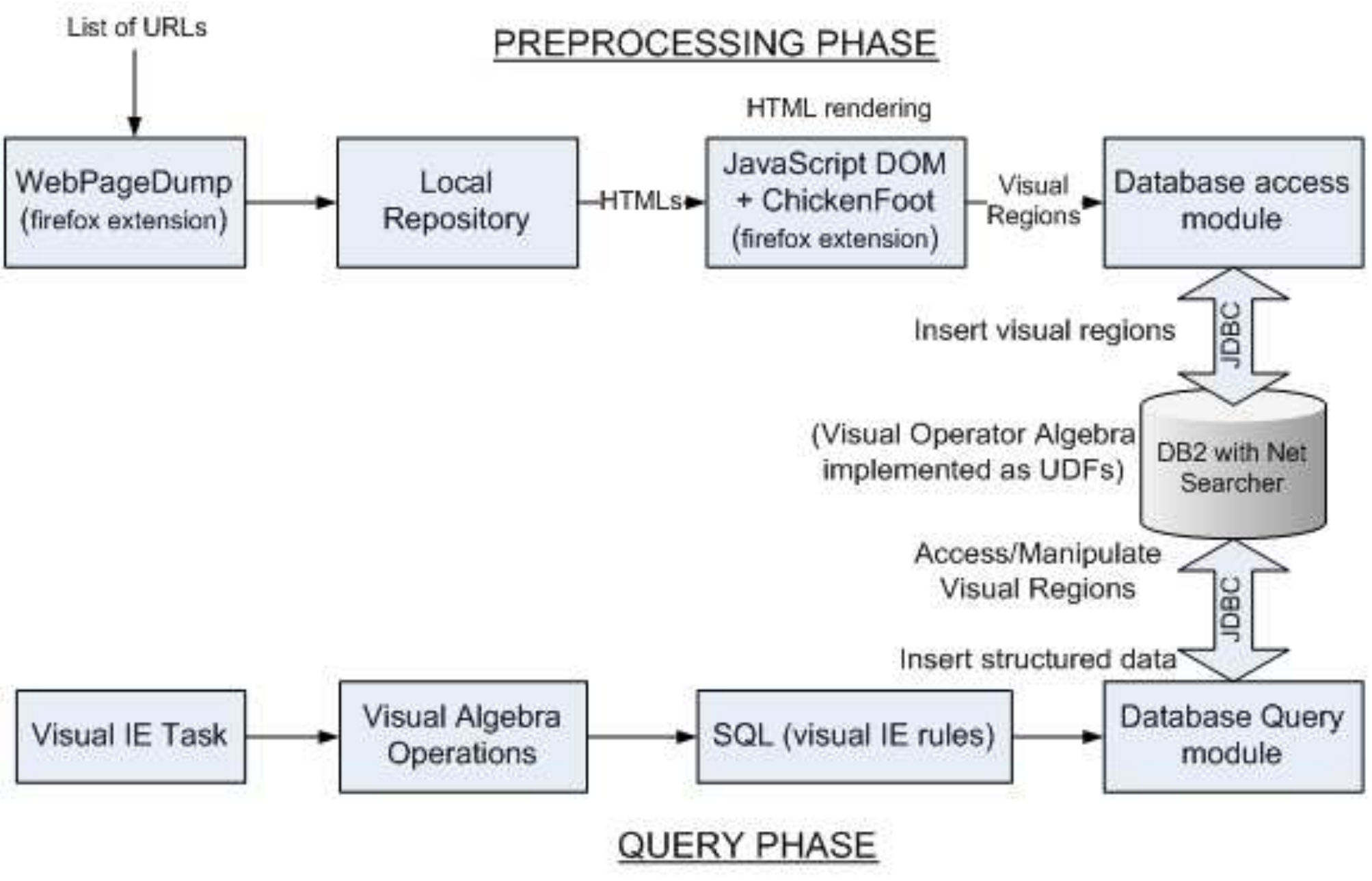}
	\caption{System Architecture}
	\label{fig:arch}
\end{figure*}

This section describes the architecture and our implementation of the visual extraction system. There are two models typically used for information extraction -- document level processing, in which rules are applied to one document at a time and collection level processing, in which the rules are matched against the entire document collection at once. 
The document at a time processing is suitable in the scenario where the document collection is dynamic and new documents are added over time. The collection level processing is useful when the document collection is static and the rules are dynamic, i.e. new rules are being developed on the same collection over time. Previous work has demonstrated an order of magnitude improvement in performance by collection level processing compared to document level processing with the use of indices for evaluating regular expression rules~\cite{ramakrishnan-balakrishnan-joshi:2006:EMNLP}. The visual algebra can be implemented using either a document level processing model or a collection level processing model. We implemented a collection level processing approach using a relational database with extensions for inverted indices on text for efficient query processing. 
Figure \ref{fig:arch} depicts the overall system architecture. 
Collection level processing has two phases: (a) Preprocessing phase comprising computations that can be done offline and (b) Query phase that includes the online computations done during interactive query time.

\subsubsection*{Preprocessing Phase}
In the preprocessing phase, web pages from which information is to be extracted are crawled and a local repository of the web pages is created. Along with the HTML source of the web page, all components that are required to render the page accurately, such as embedded images and stylesheets, should also be downloaded and appropriately linked from the local copy of the page. 
We use an open source Firefox extension called  WebPageDump\footnote{\url{http://www.dbai.tuwien.ac.at/user/pollak/webpagedump/}} specifically designed for this purpose. 
Each page is then rendered in a browser and for each node in the DOM tree, its visual region and text is extracted (using the Chickenfoot Firefox extension\footnote{\url{http://groups.csail.mit.edu/uid/chickenfoot/}}) and stored in a relational database (IBM DB2 UDB). We also use the indexing and text search capabilities of DB2 Net Search Extender\footnote{\url{http://www.ibm.com/software/data/db2/extenders/netsearch/}} to speed up queries that can benefit from an inverted index. 

\subsubsection*{Query Phase}
During the interactive query phase, the user expresses the information extraction task as operations in the visual algebra. The visual algebraic operations are then translated to standard SQL queries and executed on the database 

\subsection{Implementing Visual Algebra Queries using a Database}
\label{sec:db-implementation}

\subsubsection*{Schema}
The visual regions computed in the pre-processing stage are stored in table called $Regions$ with the following schema: \\
$<Pageid, Regionid, x_l, y_l, x_h, y_h, TextStart, TextEnd, \\ Text, HtmlTag, MinimalRegion, MaximalRegion>$.


The $Pageid$ uniquely identifies a page. The html DOM tree is a hierarchical structure where the higher level nodes comprise lower level nodes. For example, a {\tt <td>} may be nested inside a {\tt <tr>} tag, which is nested inside a {\tt <table>}, and so on. The $Regionid$ uniquely identifies a region in a page annd is a path expression that encodes the path to the corresponding node. This makes it easy to identify the parents and descendants of a region. For example, a node $1.2$ indicates a node reached by following the second child of the first child of the root node. The $x_l, y_l, x_h, y_h$ denote the coordinates of the region. The $Text$ field stores the text content of the node with $TextStart$ and $TextEnd$ indicating the offsets of the text within the document. The text content of higher level nodes is the union of the text content of all its children. However, to avoid duplication, we associate only the innermost node with the text content while storing in the $Regions$ table. The $MinimalRegion$ and $MaximalRegion$ fields are used to quickly identify a descendant or ancestor that has the identical text content as this node.

\subsubsection*{Implementation of Operators}
The visual algebra is implemented using a combination of standard SQL and User Defined Functions (UDFs). Due to space constraints, we mention the mapping of only some representative operators without going into complete detail in Table~\ref{table:sql-mapping}. For simplicity, we have shown the SQL for each operator separately. Applying these rules for a general algebra expression will produce a nested SQL statement that can be flattened out into a single SQL using the regular transformation rules for SQL sub-queries. We also experimented with using a spatial database to implement our algebra, but found that it was not very efficient. Spatial databases can handle complex geometries, but are not optimized for the simple rectangular geometries that the visual regions have. Conditions arising from simple rectangular geometries can be easily mapped to simple conditions on the region coordinate columns in a regular relational database.

\begin{table}
		\begin{tabular}{|l|l|}		
		  \hline
		  {\bf Operator} & {\bf Mapping} \\
		  \hline \hline
			$\Re$ & \parbox{7cm}{ \tt SELECT Pageid, Regionid FROM Regions }\\
			\hline
			$\epsilon_{re}(exp)$ & \parbox{7cm}{\tt SELECT Pageid, Regionid FROM Regions R \\WHERE MatchesRegex(R.Text, exp)} \\
			\hline
			$\epsilon_{d}(dict)$ & \parbox{7cm}{\tt SELECT Pageid, Regionid FROM Regions R \\WHERE MatchesDict(R.Text, dict)} \\
			\hline
			$Ancestors(v)$ & \parbox{7cm}{\tt SELECT Pageid, Regionid FROM Regions R \\WHERE IsPrefix(R.Regionid, v.RegionId)} \\
			\hline
			$StrictNorthOf(v_1, v_2)$ & \parbox{7cm}{\tt \ldots WHERE $v_1.y_h$ $\le$ $v_2.y_l$ $AND$ \\$v_1.x_l \ge v_2.x_l$ $AND$ $v_1.x_h \le v_2.x_h$} \\
			\hline
			$MinimalRegion(v)$ & \parbox{7.5cm}{\tt SELECT Pageid, MinimalRegion FROM Regions R \\WHERE R.Regionid = v.Regionid} \\			
			\hline
			$HorizontallyAligned(R)$ & \parbox{7cm}{\tt \ldots FROM R GROUP BY $R.x_l$} \\
			\hline
			$Minimal\-Bounding\-Region(V)$ & \parbox{7cm}{\tt SELECT $min(x_{l}), min(y_{l}), max(x_{h}), max(y_{h})$ FROM V} \\
			\hline
		\end{tabular}
	\caption{Mapping to SQL}
	\label{table:sql-mapping}
\end{table}

{\bf Visual span producing operators:}
The $\epsilon_{re}$ and $\epsilon_d$ operators are implemented using UDFs that implement regular expression and dictionary matching respectively. $Anscestors(v)$ and $Descendants(v)$ are implemented using the path expression in the region id of $vs$. Searching for all prefixes of the $Regionid$ returns the ancestors and searching for all extensions of $Regionid$ returns the descendants. 

{\bf Span aggregation operators:}
The span aggregation operators ($\Omega_o$, $\Omega_c$ and $\beta_v$) cannot be easily mapped to existing operators in SQL. We implement these in Java, external to the database. 

{\bf Other Visual operators:}
The scalar visual operators include the directional predicates, containment predicates, generalization/specialization operators and geometric operators. The predicates map to expressions in the $WHERE$ clause. 
The generalization/specialization predicates are implemented using the pre-computed values in the columns $MinimalRegion$ and $MaximalRegion$. 
The grouping operators map to $GROUP BY$ clause in SQL and the aggregate functions can be mapped to SQL aggregate functions in a straightforward way as shown for $Horizontally\-Aligned$ and $Minimal\-Bounding\-Region$. 

\subsubsection*{Use of Indices}
Indices can be used to speed up the text and region predicates. 
Instead of the $MatchesRegex$ UDF, we can use the $CONTAINS$ operation provided by the text index. We also build indices on $x_l, x_l, x_h, y_h$ columns to speed up visual operators. Once the visual algebra query is mapped to a SQL query, the optimizer performs the task of deciding what indices to use for the query based on cost implications. Example of a mapping is shown in Table~\ref{table:queries}.

%% file: experiments.tex
\section{Experiments}
\label{sec:experiments}
The goal of the experiments is two fold -  to demonstrate the simplicity of visual queries and to study the effectiveness of mapping the visual algebra queries to database queries. We describe the visual algebra queries for a representative set of tasks, map them to SQL queries in a database system and study the effect of indexing on the performance. 

\subsection{Experimental Setup}
The document corpus for our experiments consists of software product information pages from IBM web site \footnote{\url{http://www.ibm.com/software/products/us/en?pgel=lnav}}. 
We crawled these pages resulting in a corpus of $44726$ pages. Our goal was to extract the system requirements information for these products from their web pages (see Figure~\ref{fig:sys-req-2}). Extracting the system requirements is a challenging task since the pages are manually created and don't have a standard format. This can be broken into sub-tasks that we use as representative queries for our experiments. 
The queries are listed below. The visual algebra expression and the equivalent SQL query over the spatial database are listed in Table~\ref{table:queries}. For ease of expression, the visual algebra queries are specified using a SQL like syntax. The functions $RegEx$ and $Dict$ represent the operators $\epsilon_r$ and $\epsilon_d$ respectively. For each of these sub-tasks, it is possible to write more precise queries. However, our goal here is to show how visual queries can be used for a variety of extraction tasks without focusing too much on the precision and recall of these queries. 
\begin{list}{\labelitemi}{\leftmargin=1em}
\item Filter the navigational bar at the left edge before extracting the system requirements. \\
{\bf Q1:} Retrieve vertically aligned regions with more than $n$ regions such that the region bounding the group is contained within a virtual region $A(x_l,y_l,x_h,y_h)$. For our domain, we found that a virtual region of $A(0,90,500,\infty)$ works well.
\item Identify whether a page is systems requirements page. We use the heuristic that system requirement pages have the term ``system requirements'' mentioned near the top of the page. 
\\
{\bf Q2:} Retrieve the region in the page containing the term 'system requirements' contained in a region $A$. In this case, we use a virtual region, $A(450,0,\infty,500)$
\item To identify various operating systems that are supported, the following query can be used.\\
{\bf Q3:} Find all regions R, such that R contains one of the operating systems mentioned in a dictionary T and are to the strict south or to the strict east of a region containing the term ``Operating Systems''.
\item To find the actual system requirements for a particular operating system, the following query can be used.\\
{\bf Q4:} Find a region that contains the term ``Windows'' that occurs to the strict south of a region containing the term ``Operating Systems'' and extract a region to the strict right of such a region.
\end{list}

Due to lack of space, we show the visual algebra expression and the equivalent SQL query (Section~\ref{sec:db-implementation}) for only query Q4 in Table~\ref{table:queries}. For ease of expression, the visual algebra queries are specified using a SQL like syntax.

\begin{table}
		\begin{tabular}{|l|l|l|}
		\hline
		 Q& Visual Query & SQL Query \\
		 \hline
4 & 
\parbox[t]{5cm}{\tt 
select R3.VisualSpan \\
from RegEx(`operating system', D) as R1, RegEx('windows', D) as R2, $\Re(D)$ as R3 \\
where StrictSouthOf(R2, R1) and StrictEastOf(R3, R2) 
} &
\begin{minipage}[t]{7.5cm}
{\tt 
SELECT R3.pageid, R3.regionid \\
FROM regions R1, regions R2, regions R3 \\
WHERE r1.pageid = R2.pageid \\ AND R2.pageid = R3.pageid AND
contains(R1.text, `"Operating Systems"') = 1 \\ AND 
contains(R2.text, `"Windows"') = 1 \\ AND 
$R2.y_l \ge R1.y_h $ AND $R2.x_l \ge R1.x_l$ \\ AND $R2.x_h \le R1.x_h$\\ AND 
$R3.x_l \ge R2.x_h $ AND $R3.y_l \ge R2.y_l$ \\ AND $R3.y_h \le R2.y_h$
}
\end{minipage} \\
\hline
		\end{tabular}
	\caption{Queries}
	\label{table:queries}
\end{table}

\subsection{Accuracy of Spatial Rules}
We measured the accuracy of our spatial rules using manually annotated data from a subset of pages in our corpus. The test set for Q2 and Q4 consists of 116 manually tagged pages. 
The test set for Q1 and Q3 contains 3310 regions from 10 pages with 525 positive examples for Q1 and 23 positive examples for Q3. Please note that for Q1 and Q3 we need to manually tag each region in a page. Since there are few hundred regions in a page, we manually tagged only 10 pages. The rules were developed by looking at different patterns that occur in a random sample of the entire corpus. The results are reported in 
Figure~\ref{table:accuracy}. Since our tasks were well suited for extraction using spatial rules, we were able to obtain a high level of accuracy using relatively simple rules. 

\subsection{Performance}

\begin{figure}[ht]
\begin{minipage}[b]{0.7\linewidth}
	\centering
		\includegraphics[width=0.7\linewidth]{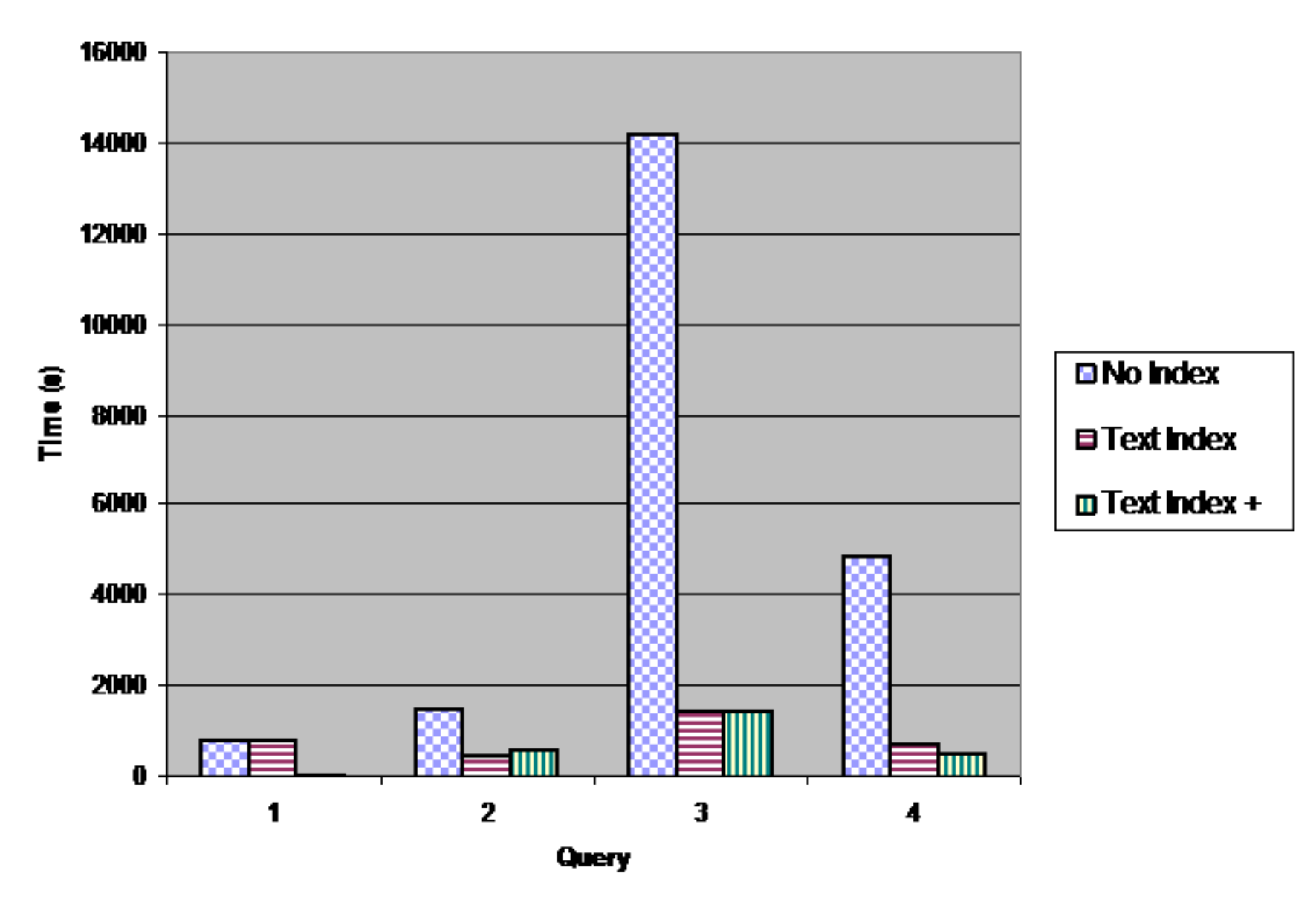}
	\caption{Effect of Indices}
	\label{fig:queryperformance}
\end{minipage}
\begin{minipage}[b]{0.2\linewidth}
{\footnotesize
		\begin{tabular}{|l|l|l|l|l|}
		  \hline
		  Query & $Q1$ & $Q2$ & $Q3$ & $Q4$ \\
		  \hline
			Recall & 100 & 96 & 100 & 100\\
			\hline
			Precision &  84  & 85 & 88 & 100 \\
			\hline
		\end{tabular}
		}
	\caption{Accuracy}
	\label{table:accuracy}
\end{minipage}
\end{figure}

We measured the performance of these queries on the document collection. Since the queries have selection predicates on the text column and the coordinates ($x_l, y_l, x_h, y_h$), we build indices to speed them up. We also index the text column using DB2 Net search extender. 
The running time for the queries are shown in Figure \ref{fig:queryperformance}. We compare various options of using no indices, using only text index and using both text index and indices on the region coordinates. For $Q1$, the text index does not make a difference since there is no text predicate. The region index leads to big improvement in the time.  $Q2$, $Q3$ and $Q4$ have both text and region predicates and thus benefit from the text index as well as the region indices. The benefit of the text index is found to be compartively larger. 
In all the cases, we can see that using indices leads to a three to fifteen times improvement in the query execution times.

%% file: conclusions.tex
\section{Discussion}
\label{sec:conclusions}
We have demonstrated an extension to the traditional rule based IE framework that allows the user to specify layout based rules. This framework can be used for many information extraction tasks that require spatial analysis without having to use custom code. The WYSIWYE algebra we propose allows the user to seamlessly combine traditional text based rules with high level rules based on spatial layout. 
The visual algebra can be systematically mapped to SQL statements, thus enabling optimization by the database. 
We have evaluated our system in terms of usability and performance for a task of extracting software system requirements from software web pages. The rules expressed using the visual algebra are much simpler than the corresponding source based rules and more robust to changes in the source code. 
The performance results show that by mapping the queries to SQL and using text and region indexes in the database, we can get significant improvement in the time required to apply the rules. 

Layout based rules are useful for certain types of pages, where the layout information provides cues on the information to extract. A significant source of variation in web pages (different source code, same visual layout) can be addressed by rule based information extraction systems based on a visual algebra, leading to simpler rules. Visual rules are not always a replacement for the text based rules, rather they are complementary. In our system, we can write rules that combine both text based and layout based rules in one general framework.